\documentclass{article}

\PassOptionsToPackage{numbers, compress}{natbib}

\usepackage[preprint]{neurips_2020}

\usepackage[utf8]{inputenc} 
\usepackage[T1]{fontenc}    
\usepackage{hyperref}       
\usepackage{url}            
\usepackage{booktabs}       
\usepackage{amsfonts}       
\usepackage{nicefrac}       
\usepackage{microtype}      

\usepackage{backref}
\usepackage{xcolor}
\usepackage{nicefrac}
\usepackage{graphicx}
\usepackage{caption}
\usepackage{subcaption}
\usepackage{algorithm}
\usepackage{algorithmic}
\usepackage{amsmath}
\usepackage{mathtools}
\usepackage{comment}
\usepackage{pifont}

\hypersetup{
    colorlinks,
    linkcolor={red},
    citecolor={red},
    urlcolor={blue}
}

\title{Insights from the Future for Continual Learning}

\author{%
  Arthur Douillard\thanks{Corresponding author at \texttt{arthur.douillard@heuritech.com}} \\
  Sorbonne Université\\
  Heuritech\\
  \And
  Eduardo Valle \\
  University  of Campinas \\
  \And
  Charles Ollion\\
  Heuritech\\
  \And
  Thomas Robert\\
  Heuritech\\
  \And
  Matthieu Cord\\
  Sorbonne université\\
  valeo.ai
}

\definecolor{BrickRed}{HTML}{B6321C}
\definecolor{RoyalBlue}{HTML}{0071BC}
\definecolor{PineGreen}{HTML}{008B72}
\definecolor{bluefig}{HTML}{5B9BD5}
\definecolor{Gray}{gray}{0.9}

\newcommand{\arthur}[1]{\textcolor{RoyalBlue}{Arthur: #1}}

\newcommand{\eduardo}[1]{\textcolor{BrickRed}{Eduardo: #1}}

\newcommand{\gray}[1]{{\color{darkgray}#1}}

\let\thetaold\theta
\renewcommand{\theta}{\boldsymbol{\thetaold}}
\newcommand{\mcL}{\mathcal{L}}
\newcommand{\vx}{\mathbf{x}}
\newcommand{\vh}{\mathbf{h}}
\newcommand{\vy}{\mathbf{y}}
\newcommand{\vt}{\mathbf{\theta}}
\newcommand{\vyh}{\hat\vy}

\newcommand{\std}{$\pm\,$}
\newcommand{\clf}{\textit{clf}}
\newcommand{\tableindent}{\,\,\,\,}

\begin{document}

\maketitle

\begin{abstract}
Continual learning aims to learn tasks sequentially, with (often severe) constraints on the storage of old learning samples, without suffering from catastrophic forgetting. In this work, we propose prescient continual learning, a novel experimental setting, to incorporate existing information about the classes, prior to any training data. Usually, each task in a traditional continual learning setting evaluates the model on present and past classes, the latter with a limited number of training samples. Our setting adds future classes, with no training samples at all. We introduce Ghost Model, a representation-learning-based model for continual learning using ideas from zero-shot learning. A generative model of the representation space in concert with a careful adjustment of the losses allows us to exploit insights from future classes to constraint the spatial arrangement of the past and current classes. Quantitative results on the AwA2 and aP\&Y datasets and detailed visualizations showcase the interest of this new setting and the method we propose to address it.
\end{abstract}



\section{Introduction}
\label{sec:introduction}


Continual learning models contrast with traditional models by approaching a sequence of tasks incrementally. With limitations (often severe) on the training data they can retain, those models must avoid catastrophic forgetting of past tasks \cite{robins1995catastrophicforgetting, french1999catastrophicforgetting}, while remaining receptive to new tasks. Many approaches exist to counteract forgetting: keeping a limited amount of training data from previous tasks~\cite{rebuffi2017icarl,castro2018end_to_end_inc_learn}; learning to generate the training data~\cite{kemker2018fearnet,shin2017deep_generative_replay}; extending the architecture for new tasks~\cite{yoon2018dynamically_expandable_networks,li2019learning_to_grow}; using a sub-network for each task~\cite{fernando2017path_net,golkar2019neural_pruning, hung2019cpg}; and constraining the model divergence as it evolves~\cite{kirkpatrick2017ewc,lopezpaz2017gem,aljundi2018MemoryAwareSynapses,li2018lwf,rebuffi2017icarl,castro2018end_to_end_inc_learn, douillard2020podnet}. Those approaches are often complementary.

We propose a challenging new setting, \textit{prescient continual learning}, in which the model must perform well not only for present and past tasks, but also for \textit{future} ones, both avoiding catastrophic forgetting (using a limited number of training samples for past classes), and giving the best possible estimates for the future classes (using no training samples at all). To make the setting possible, the model must know the classes and have some prior information about them.
Indeed, Aljundi et al. \cite{aljundi2019selfless} remark that the ability to make room for future classes is a key limitation of current continual learning models, and propose a regularization loss to make the model more “selfless”, explicitly leaving capacity for future classes in the representation. Han et al. \cite{hanrebuffi2020autodiscovering} proposed a setting where the training samples from all classes are present from the beginning, but the labels become available incrementally. In a way, our setting is the inverse: we know which labels we are going to encounter, but the training data for those labels arrive incrementally. For instance, in many real-world applications (e.g. fashion product classification), due to budget constraints, models are released incrementally, with partial classes and training data, despite the classes being known from the beginning, and being well-characterized by attributes.

To address the proposed setting, we take inspiration from zero-shot learning
\cite{lampert2009zeroshot, xian2019awa2}, which allows classifying examples from unseen classes by combining a vision model with an embedding possessing knowledge about the classes (e.g. a word embedding \cite{mikolov2013word2vec,pennington2014glove} or an attributes matrix). Although several approaches exist for zero-shot learning, we will focus on generating a representation for the future classes \cite{bucher2017zeroshot_gmmn, kumar2018synthesized_zeroshot, xian2018feature_generating_zeroshot}. 
The framework of representation learning will allow us to integrate continual and zero-shot learning seamlessly, as we advance through the tasks and future classes become present classes, and then past classes. Moreover, we will be able to use \textit{ghost features}, predicted features for the future classes, to make room in the representation space for future classes. All those goals all integrated into a simple, streamlined model due to a careful construction of the losses.

The contributions of this work are two-fold: (1) we propose a new challenging setting, \textit{prescient} continual learning, where the model must perform well on past, present, and \textit{future} classes; (2) we propose our \textit{ghost model} to address that setting, integrating continual and zero-shot learning into a coherent whole. We evaluate those contributions in comprehensive experiments, showcasing both the intuitive appeal of our model, and its performance.

\section{Setting: prescient continual learning}

\label{sec:setting}
\begin{figure}
\centering
  \includegraphics[width=0.7\linewidth]{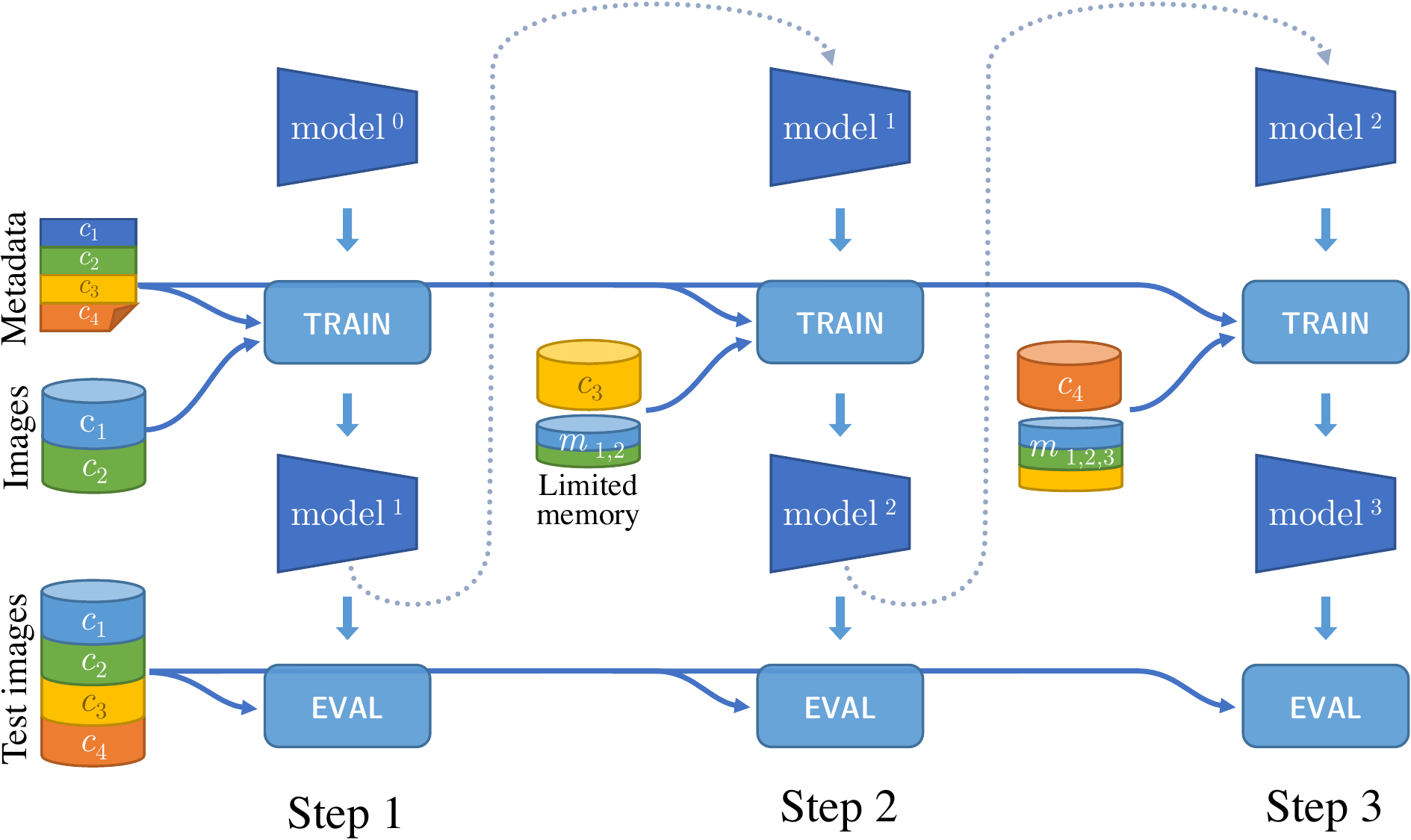} 
    \caption{The enriched continual learning setting proposed in this work. At each training task, we learn a new set of classes, but the model is evaluated on \textit{all classes} --- past, present, and future. The model has to avoid catastrophic forgetting of past classes (using a limited number of rehearsal training samples), as well as make a good guess for future classes (using no training samples at all).}
\label{fig:protocol_zeroshot_continual}
\end{figure}

In continual learning, a classifier is trained in multiple steps called \textit{tasks}. Each task $t\in{1:T}$ comprises a set of new classes $C^t$. The model is evaluated after each task $t$, traditionally, on all classes seen so far $C^{1:t}$. Our \textit{rehearsal memory} severely limits the training samples we are allowed to keep from the past classes: following \cite{hou2019ucir,douillard2020podnet}, we allow a small constant number of samples $s$ per past class. We must take the maximum advantage from those limited rehearsal data to avoid catastrophic forgetting. 

We propose an enriched experimental setting, prescient continual learning, in which each task is evaluated on \textit{all classes} $C^{1:T}$: past ($C^{1:t-1}$), present ($C^t$), and future ($C^{t+1:T}$). In that challenging new setting, we must not only avoid the catastrophic forgetting of past classes (using the limited rehearsal training samples), but also give our best estimates for future classes (using no training samples at all). That will only be possible if we have some prior information about the classes, e.g., their hierarchy in a semantic network (like WordNet), an associated word embedding (like Word2Vec), or an attribute matrix. Such setting is illustrated in \autoref{fig:protocol_zeroshot_continual}. We will sometimes shorthand the set of past and present classes $C^{1:t}$ as the \textit{seen} classes, and the set of future classes $C^{t+1:T}$ as the \textit{unseen} classes. We denote individual samples by a superscript $\vx^(i)$, and the class label by a subscript $\vx_c$. We denote on which parameters a loss is applied by a subscript $\mcL_\Theta$.

\section{Ghost model}
\label{sec:model}

To address the setting described in the previous section, we propose our \textit{ghost model}, comprising three components: a convolutional feature extractor $f$, a feature generator $g$, and a classifier $\clf$. The feature extractor is the backbone of the model: it learns to extract a feature vector from actual samples that can be fed to the classifier. The generator learns the distribution of the features for all classes, aiming to generate plausible samples of features for the future classes. The classifier makes the final decision for all classes: past, present, and future. The classifier is trained on future classes with features sampled from the generator, which we call \textit{ghost features} (since they must be “hallucinated” from the seen classes and some prior knowledge about the classes).

All three components are learned continuously throughout the $T$ tasks; we denote the component learned at task $t$ with a superscript: $f^t$, $g^t$, and $\clf^t$. To avoid catastrophic forgetting of past classes, we constrain the model's evolution, placing a cost on the divergence between $f^{t-1}$ and $f^t$. To make the best guess for future classes, we take inspiration from zero-shot learning, training the classifier with the ghost features. Finally, we adjust the classifier so that the ghost features “make room” in the representation space for future classes.

The next subsections detail the model, explaining the different terms in the loss that act in concert to obtain the desired balance. 

\subsection{Base model for continual learning}
\label{sec:basemodel}

The base model is a representation-based architecture with a convolutional feature extractor $\vh = f(\vx)$ (where $\vx$ is the input image and $\vh$ is the feature vector) and a cosine classifier $\clf$ \cite{luo2018cosine_classifier, hou2019ucir}, a fully-connected layer with the dot-product replaced by a cosine similarity:
\begin{equation}
    \clf(\vh, \vt)_c = \vyh_c = \frac{\langle \vh, \vt_c \rangle}{\Vert \vh \Vert_2 \Vert \vt_i \Vert_2}\,.
    \label{eq:cosine-classifier}
\end{equation}
Remark that $\vt_c$, the parameter vector for the $c$-th class in $\clf$ ($c\in C^{1:t}$), may be interpreted as a representative or proxy for that class. The classification loss could be either a cross-entropy loss preceded by softmax activations, or an NCA loss \cite{goldberger2005nca_loss, attias2017proxynca, douillard2020podnet}:
\begin{equation}
    \mcL^{\text{\tiny{nca}}}_{\Theta_f,\Theta_{\clf}} = \left[- \log\frac{\exp\left(\vyh_y- \delta\right)}{\sum_{c \neq y} \exp \vyh_c} \right]_+ \,.
    \label{eq:nca}
\end{equation}
To counteract catastrophic forgetting, we must limit the evolution of the model. We impose — as usual for continual learning — a distillation loss between the previous model iteration ($t-1$) and the current one ($t$). We evaluate several distillation losses applied to intermediate and final outputs of the feature extractor \cite{douillard2020podnet,hou2019ucir} on \autoref{sec:implementation}. The final loss of the base model is:
\begin{equation}
    \mcL = \mcL^{\text{\tiny{nca}}}_{\Theta_f,\Theta_{\clf}} + \lambda_1 \mcL^\text{\tiny{distil}}_{\Theta_f}\,.
    \label{eq:basemodel_loss}
\end{equation}

\subsection{Capacitating ghost model for future classes}

The base model can deal with both present classes (with training samples constrained only by their availability in the training set)  and past classes (with training samples severely constrained by the rehearsal memory). As discussed, the introduction of a distillation loss prevents catastrophic forgetting of the latter. We will now address \textit{future classes}, with \textit{no} training samples available. First, taking inspiration from zero-shot learning, we will use prior information about the classes to generate \textit{ghost features}, plausible stand-ins for the unseen future classes' features. Next, we will adapt the classifier to incorporate those ghost features into the learning objective seamlessly. The representation learning framework will allow us to integrate the entire learning apparatus into one coherent loss.

\textbf{Generator.~~} The generative model estimates the distribution of the unseen classes directly in terms of their features (instead of the input images). For the feature generation to work, we must have exploitable prior information about the classes, more precisely, we must be able to map the class labels $c$ into a \textit{class attribute space} that makes semantic sense. The exact way to perform that mapping will be data-dependent, but most often, either we will have an explicit set of attributes linked to each class (color, size, material, provenance, etc.), or we will be able to extract a latent semantic vector, using a technique like Word2vec \cite{mikolov2013word2vec,pennington2014glove}. The generator learns to link the attributes of the \textit{seen} classes to the actual feature vectors extracted from the training samples of those classes. Thus the first generator training must happen after the features extractor (its ground-truth) is learned. The generator is fine-tuned after each task to handle distribution shift. Next, we ask the generator to draw random samples, using the attributes of the \textit{unseen} classes, creating counterfeit features that we call \textit{ghost features}. The strategy is agnostic to the generator model as long as it can be conditioned by class attributes.  At present, as detailed in \autoref{sec:quantitative}, we choose a Generative Moment Matching Network \cite{li2015gmmn}: a shallow multi-layer perceptron conditioned by class attributes and a noise vector trained to minimize the Maximum Mean Discrepancy \cite{gretton2007twosampleMMD,gretton2012twosampletestMMD}.

\label{sec:generator}

\textbf{Complete classifier.~~} Remind that the parameters  $\{\vt_c\,,\forall c \in C^{1:t}\}$ on the representation-based classifier (\autoref{eq:cosine-classifier}) may be interpreted as proxies for the classes $C^{1:t}$. The base model for task $t$ will, thus, learn $|C^{1:t}|$ such proxies, one for each of the seen classes. To extend the model for the unseen future classes, the complete classifier will learn $|C^{1:t}| + |C^{t+1:T}|$ proxies, which changes \autoref{eq:nca} to:
\begin{equation}
    \mcL^\text{\tiny{nca-ghost}}_{\Theta_f,\Theta_{\clf}} = \left[- \log\frac{\exp\left(\vyh_y - \delta\right)}{\sum_{\substack{c \neq y\\c \in C^{1:t}}} \exp \vyh_{c} + \sum_{\substack{c \neq y\\c \in C^{t+1:T}}} \exp \vyh_{c}} \right]_+ \,.
    \label{eq:nca-ghost}
\end{equation}
This classification loss maximizes the similarity $\vyh_y$ (or, conversely, minimizes the distance) between sample feature $\vh_y$ and correct class proxy $\vt_y$ in the numerator. In the denominator, the loss pushes away all wrong class proxies, from both seen and unseen (future) classes, by minimizing the similarities with $\vy_c, \, \forall c \neq y$.

The participation of future classes in the classification loss has two effects. Most obviously, it allows the model to perform zero-shot-like guesses for those classes during test time. The representation-learning paradigm allows performing both continual and zero-shot learning seamlessly, as we advance through the tasks and future classes become present classes, and then past classes. Less evidently, but vitally important, the learning of proxies for the future classes makes room in the representation space for those classes, creating effective empty spaces that push away the actual features of the seen classes (due to the repulsive term in the denominator). As we advance through training, future classes become present, their ghost placeholders disappear, and they can neatly fit in the newly vacant region. Such a strategy reduces interference between classes throughout continual learning, and, as we will see in both visual and quantitative experiments, has long-range positive effects.

Naturally, the complete classifier has to be trained with samples from all classes. For seen classes, actual data is available from the training and rehearsal data. For unseen future classes data is not available, so we employ ghost features sampled from the generator. Note that ghost features are produced once per task by the generator and are kept fixed for the task duration.
\label{sec:ghost-classifier}


\label{sec:svm}
\textbf{Latent-space regularization.~~} 
As explained above, our Ghost classification loss minimizes the intra-class distances and maximizes the inter-class distances. The loss enforces those constraints to all proxies regardless of whether they represent seen classes or not. We further promote an inter-class separation by optimizing the latent representation of seen classes to avoid overlapping with the representation of Ghosts. That loss constrains the features space directly and does not affect the proxies and the intra-class distances.

\begin{figure}
\centering
  \includegraphics[width=0.7\linewidth]{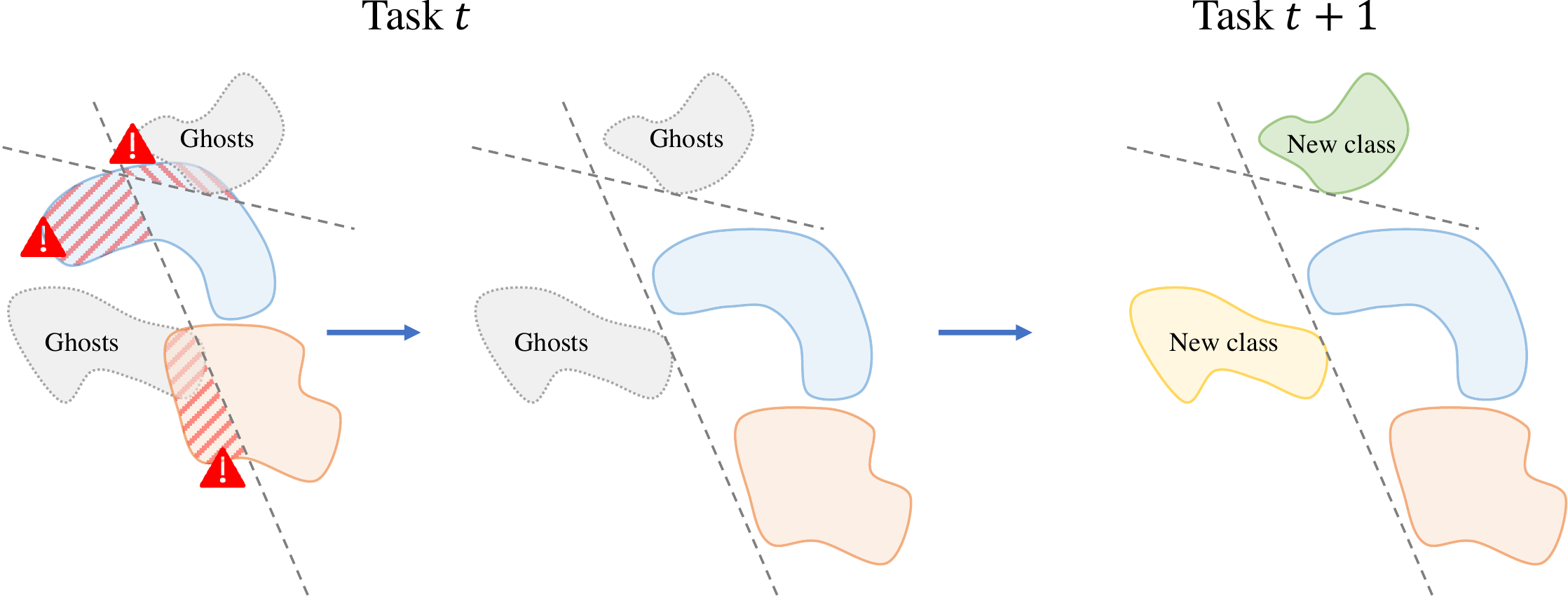}
\caption{Latent-space regularization establishes margin-based one-unseen-class vs. all-seen-classes linear separations. Those separations are employed to directly condition the feature space, creating space for future unseen classes. In the following task, some unseen class will become seen, and may occupy the feature space with less interference.}
\label{fig:svm_reg}
\end{figure}

We based this regularization loss on SVM \cite{cortes1995svm} for simplicity, but other methods could have similar behavior. To compute that loss, we learn binary one-unseen-class-Vs-all-seen-classes SVM classifiers, one for each unseen class. We employ a linear kernel, since the feature extractor and feature vector dimensionality (512) allows good linear separation, but more complex kernels could be used. Those SVMs define hyperplanes $\mathbf{w}_c$ and biases $b_c,\, \forall c \in C^{t+1:T}$, separating each unseen region from the mass of seen features $\mathbf{h}^{(i)}$ (\autoref{fig:svm_reg}):
\begin{equation}
    \mcL^\text{\tiny{svm-reg}}_{\Theta_f} = \frac{1} {N\times |C^{t+1:T}|}\sum_{i=1}^N \sum_{c\in C^{t+1:T}} [\mathbf{w}_c \cdot \mathbf{h}^{(i)} + b_c + \tau]_+\,,
    \label{eq:svm}
\end{equation}
where $\mathbf{h}^{(i)}$ are seen features (classes in $C^{1:t}$), $[\,\cdot\,]_+$ the hinge loss, and $\tau$ an additional margin (higher values of $\tau$ push seen features further away from the ghost regions, in practice, we set $\tau=1$ to repel beyond the support vectors).

The margin-based regularization of \autoref{eq:svm} refines the ghost classification loss of \autoref{eq:nca-ghost}. While the latter acts over the classifier conditioning the feature space indirectly via the action of the class proxies, the former acts directly over the latent/feature space and the feature extractor backbone that creates it. The computational overhead of training several SVMs, detailed in the supplementary material, is negligible compared to the total training time.

\textbf{Complete strategy.~~} All modules and losses fit neatly into the goal of learning continuously over seen and unseen classes. We train feature extractor (plus classifier) and generator in alternation. We train the latter to mimic the features of seen classes, and then ask it to extrapolate to unseen classes (ghost features). Ghost features allow us both to unify seen and unseen classes into a complete classifier ($\mcL^\text{\tiny{nca-ghost}}$), and to enforce early allocation in the feature space for unseen classes ($\mcL^\text{\tiny{svm-reg}}$). The complete loss, in addition to a distillation loss to counter-act catastrophic forgetting ($\mcL^\text{\tiny{distill}}$), is:
\begin{equation}
    \mcL = \mcL^\text{\tiny{nca-ghost}}_{\Theta_f,\Theta_{\clf}} + \lambda_1 \mcL^\text{\tiny{distill}}_{\Theta_f} + \lambda_2 \mcL^\text{\tiny{svm-reg}}_{\Theta_f}\,.
\end{equation}
An algorithm describing the model's training is provided in the supplementary material.

\section{Experiments}
\label{sec:implementation}

\subsection{Pictorial experiments}

Before running our full-scale experiments, we will perform a set of experiments with a model that keeps the main components from the proposed model in a simplified form that will allow us to link quantitative performances to intuitive visual plots of the feature spaces. For the experiments in this section we employ the MNIST \cite{lecun2010mnist} dataset, with an initial task of 6 classes (digits 0 to 5), then two more tasks of two classes each; the feature extractor comprises two convolutional layers followed by a fully connected layer outputting a feature vector of only two dimensions — a purposeful choice to allow easy visualization of the feature space. Because MNIST classes have no attributes, we cannot apply zero-shot learning directly. Instead, we employ the features of actual images from the future classes instead of samples from the generator — which corresponds, in some ways, to having a perfectly calibrated generator. Those features are extracted once per task, and the feature extractor is never trained on unseen classes images. As we will see in \autoref{tab:generated_vs_real}, integrating the generator in our full-scale model outperforms using actual extracted features, so that necessary substitution does not exaggerate the abilities of this small-scale model. The losses used to train the small and the full-scale models are the same, but the SVM-based regularization was not employed since it made little sense for a 2D latent space.

\begin{figure}
\centering
  \includegraphics[width=0.8\linewidth]{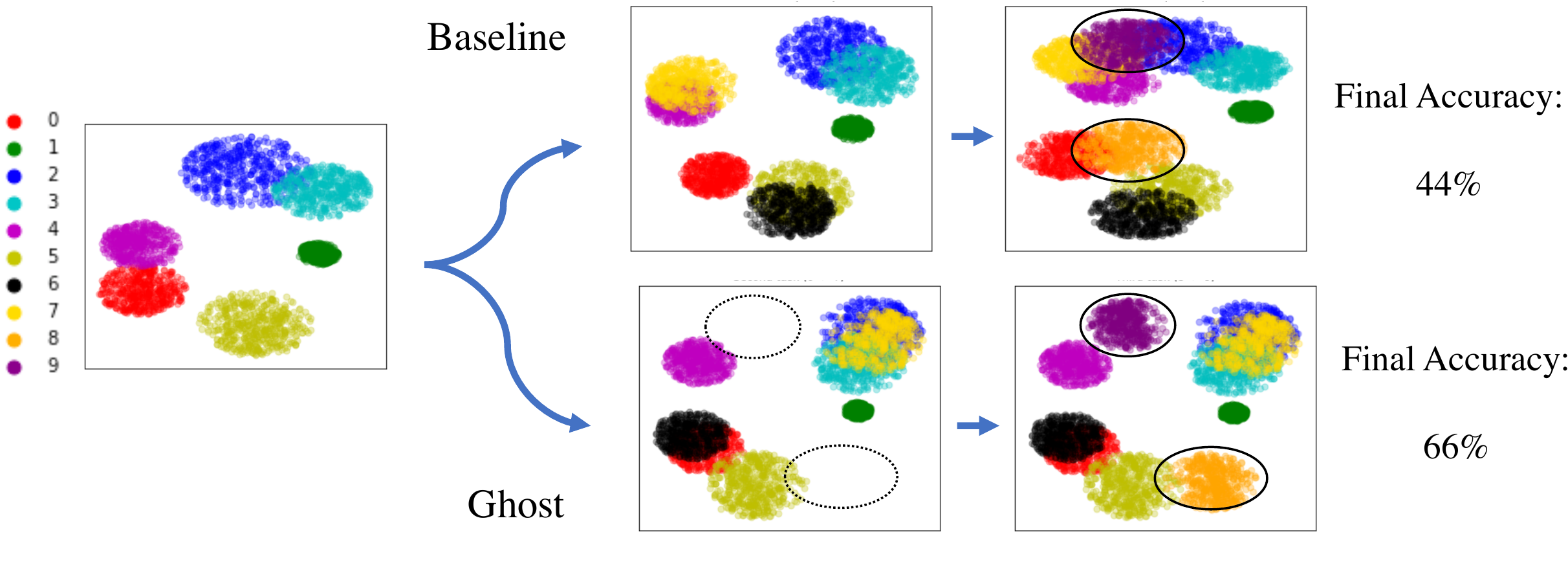}
\caption{Small-scale PODNet on MNIST with 3 steps (digits '0' to '5'; then '6' and '7'; then  '8' and '9') with a features space of only two dimensions. The early incorporation of ghost features/proxies in the second task, denoted by dotted circles in the bottom row, enforces vacant space for those unseen classes. When filled in the third task (last column), there is less interference/overlap with previous classes. Such a strategy improves the final accuracy by 22 p.p.}
\label{fig:toy_ghost_weights_3steps}
\end{figure}

The 2D feature space allows us to directly visualize the evolution of the feature space as the tasks progress, without the need for dimensionality reduction techniques that complicate the analysis (e.g. t-SNE \cite{maaten2008tsne}). \autoref{fig:toy_ghost_weights_3steps} may be interpreted upfront: as the three tasks progress left-to-right, we see the evolution of the feature space on the base model (PODNet, on the top branch) and on the proposed model with ghost features (bottom branch). The base model presents strong overlap between the initial classes, and the later added  8 ({\color{orange}orange}) and 9 ({\color{violet}dark purple}). That comes partially from shape similarities (‘8’ is similar to ‘0’ and ‘5’), partially from continual learning, and results in severe forgetting of old classes in favor of new ones. The proposed model organizes better the feature space, which is particularly visible between the second and third steps, where the early allocation of ghost zones for the future classes (displayed as empty black circles) is prominent. That better arrangement of the feature space increases the final accuracy from 44 to 66\%, a 22 p.p. improvement. The small latent space of 2 dimensions explains the low performance for both models; a model that learns on all classes in one step (\textit{i.e.} not in a continual setting) reaches only 69\% of final accuracy. That said, our model shows a clear improvement both qualitatively in \autoref{fig:toy_ghost_weights_3steps} and quantitatively. More experiments of this kind can be found in the supplementary material.


\subsection{Main experiments}
\label{sec:quantitative}

\textbf{Datasets \& Protocols.~~} We perform our experiments on two datasets: AwA2 \cite{xian2019awa2}, with 50 animals categories, each with 85 attributes; and AP\&Y \cite{farhadi2009apy}, with 32 everyday object classes, each with 64 attributes. 
We employ two experimental protocols: one typical for continual learning, following \cite{hou2019ucir,douillard2020podnet}, starting the first task with half the classes (i.e., 25 for AwA2, and 16 for aP\&Y), then adding the remaining classes in evenly-sized tasks (5 tasks of 5 classes for AwA2, and 8 tasks of 2 classes for aP\&Y); another inspired from zero-shot learning, following \cite{xian2019awa2}, starting with a standard selection of classes for each of the datasets (40 for AwA2, and 20 for aP\&Y), and adding the remaining classes in small increments (5 tasks of 2 classes for AwA2, and 6 tasks of 2 classes for aP\&Y). Our evaluation protocol is akin to the challenging and realist Generalized Zero-shot Learning \cite{scheirer2013generalizedzeroshot, chao2016generalizedzeroshot} protocol  — with no information on whether a sample is from a seen or unseen class — but harsher, since classes are seen gradually, and training samples for past classes data are limited by rehearsal memory.

\textbf{Base Models.~~} We evaluate our contributions on top of two different representation-learning based models, both based on ResNet18 \cite{he2016resnet} feature extractor backbones (with feature vector size of 512) and cosine classifiers. They differ on the distillation loss $\mcL_\text{distil}$ employed, the first model (PODNet) using Douillard et al.'s distillation \cite{douillard2020podnet} constraining the statistics of the intermediate features after each residual block, and the second (UCIR) using Hou et al.'s distillation \cite{hou2019ucir} enforcing a cosine constraint on the final flat latent space. The former performs better than the latter, but both are improved by the innovations proposed in this work. The cosine classifier has a single proxy/representative per class but could easily be generalized to multiple proxies. Implementation details are provided in the supplementary material and the code for the entire model will be released with the final paper.

\textbf{Generator.~~} Following the work of Bucher et al. \cite{bucher2017zeroshot_gmmn} for zero-shot learning, our generator is a Generative Moment Matching Network (GMMN)~\cite{li2015gmmn} $g^t(\mathbf{\xi}, \mathbf{E}_c)$, which takes as inputs a Gaussian noise vector $\mathbf{\xi}$ and a class attributes vector $\mathbf{E}_c$, and outputs a sample from the estimated distribution of features for a class with the given attributes. In our experiments on AwA2 and AP\&Y, the class attributes vectors are the average of the attributes for the training samples in the class. For each task $t$, the feature extractor $f^t$ and the generator $g^t$ are trained to minimize the Maximum Mean Discrepancy (MMD) \cite{gretton2007twosampleMMD,gretton2012twosampletestMMD} between the actual features of seen classes $\vh_c = f^t(\vx_c) \, , \forall c \in C^{1:t}$ and their distribution on the generator $\tilde{\vh}_c = g^t(\mathbf{\xi}, \mathbf{E}_{c}) \,, \forall c \in C^{1:t}$. More details and figures can be found in the supplementary material.

\begin{table*}
\caption{Continual Accuracy on AwA2 and aP\&Y for PODNet and UCIR.}
\label{tab:continual_half}
\centering
\begin{tabular}{@{}l|cc|cc@{}}
 \toprule
 &  \multicolumn{2}{c}{AwA2} &  \multicolumn{2}{c}{aP\&Y}\\
 &  \multicolumn{2}{c}{25 classes + 5 $\times$ 5 classes} &  \multicolumn{2}{c}{16 classes + 8 $\times$ 2 classes} \\
 & PODNet \cite{douillard2020podnet} &  UCIR \cite{hou2019ucir} & PODNet \cite{douillard2020podnet} &  UCIR \cite{hou2019ucir}\\
\midrule
Baseline & 62.92 \std 0.12 & 54.80 \std 0.40 & 58.64 \std 0.66 & 43.42 \std 0.21 \\
\tableindent+ $\mcL^\text{\tiny{nca-ghost}}$     & 68.31 \std 0.36 & 57.88 \std 0.27 & 62.08 \std 0.25 & 50.23 \std 0.29 \\
\tableindent+ $\mcL^\text{\tiny{nca-ghost}}$ + $\mcL^\text{\tiny{svm-reg}}$    & 68.46 \std 0.47  & 58.08 \std 0.46 & 62.73 \std 0.60 & 50.91 \std 0.56\\
\bottomrule
\end{tabular}
\end{table*}
\begin{table*}
\caption{Final Accuracy on AwA2 and aP\&Y for PODNet and UCIR.}
\label{tab:final_half}
\centering
\begin{tabular}{@{}l|cc|cc@{}}
 \toprule
 &  \multicolumn{2}{c}{AwA2} &  \multicolumn{2}{c}{aP\&Y}\\
 &  \multicolumn{2}{c}{25 classes + 5 $\times$ 5 classes} &  \multicolumn{2}{c}{16 classes + 8 $\times$ 2 classes} \\
 & PODNet \cite{douillard2020podnet} &  UCIR \cite{hou2019ucir} & PODNet \cite{douillard2020podnet} &  UCIR \cite{hou2019ucir}\\
\midrule
Baseline & 77.63 \std 0.06 & 67.07 \std 0.81 & 57.80 \std 0.97 & 42.23 \std 1.34 \\
\tableindent+ $\mcL^\text{\tiny{nca-ghost}}$ & 78.70 \std 0.46 & 67.43 \std 0.08 & 62.47 \std 0.40 & 44.17 \std 1.48\\
\tableindent+ $\mcL^\text{\tiny{nca-ghost}}$ + $\mcL^\text{\tiny{svm-reg}}$ & 79.08 \std 0.53 & 67.53 \std 0.45 & 63.30 \std 0.98 & 45.97 \std 0.26\\
\bottomrule
\end{tabular}
\end{table*}

\textbf{Continual Learning with Future Classes.~~} 
For continual learning, it is usual to take into account the model's performance as it evolves. We adapt the traditional average incremental accuracy \cite{rebuffi2017icarl} to take into account \textit{all} classes, including the future ones, and call that metric \textit{continual accuracy}: the average of accuracy over all seen classes after each task. The results appear in \autoref{tab:continual_half}, which shows, for the datasets and protocols explained in the top of this section, the performance for our two base models \cite{hou2019ucir,douillard2020podnet}, and the improvements on those base models as we implement our proposed model, with and without the SVM latent-space regularization refinement. The ability to guess on future class brings large improvements on both datasets, for both base models. The SVM-based regularization refinement also improves the results, by up to 0.65 p.p.

\textbf{Final Accuracy.~~} Once we reach the final task, the proposed model ability to guess future classes provides no advantage due to all classes being now seen. Still, as shown in \autoref{tab:final_half} — where the metric is simply the accuracy at the final task of each run — the proposed method outperforms the baselines, due to a better organization of the feature space. Although the numerical advantages in this table are smaller than in the previous one, these results are consequential, showing that the ability of the proposed model of incorporating knowledge about the classes is useful beyond the zero-shot scenario. Again, the SVM-regularization refinement helps by up to 1.80 p.p.

\begin{figure}
\centering
\begin{subfigure}{0.35\linewidth}
  \centering
  \includegraphics[width=\linewidth]{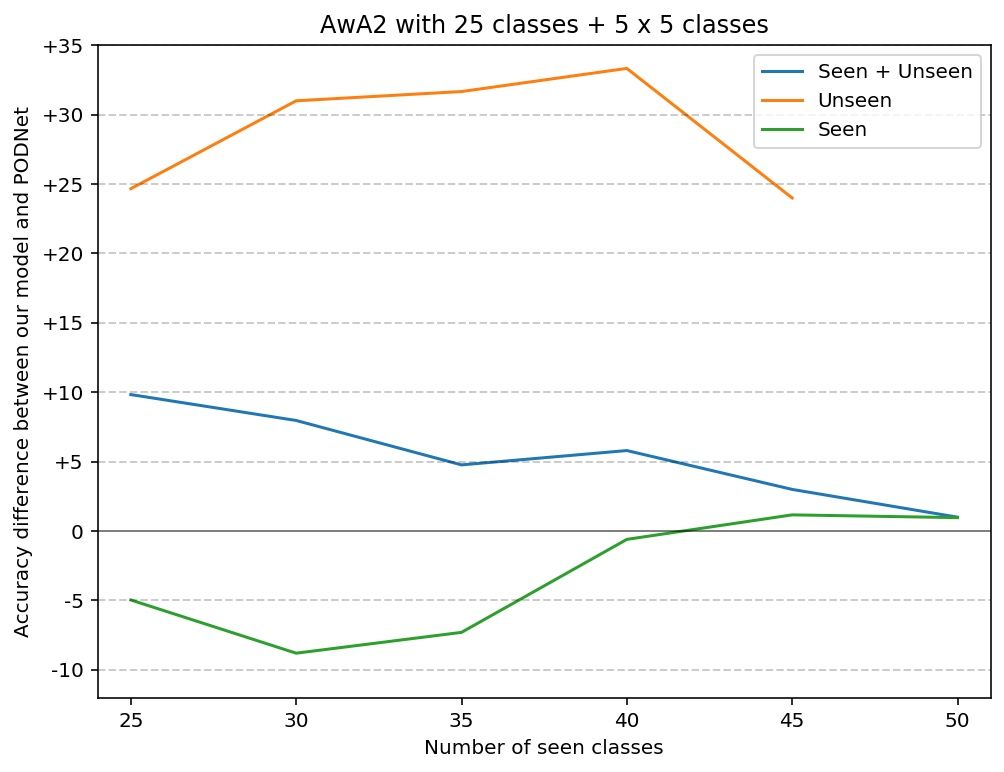}
  \caption{Ghost vs PODNet \cite{douillard2020podnet}.}  
\end{subfigure}
\begin{subfigure}{0.35\linewidth}
  \centering
  \includegraphics[width=\linewidth]{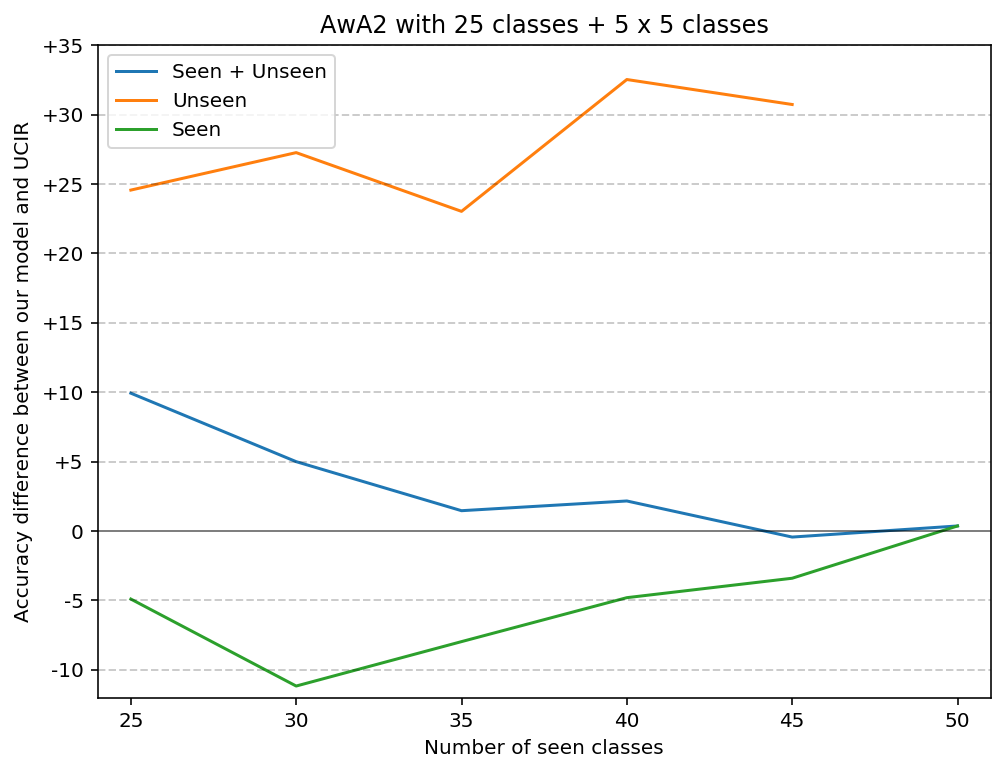}
  \caption{Ghost vs UCIR \cite{hou2019ucir}}
\end{subfigure}
\caption{Ghost model vs base models on AwA2, difference of accuracy over all classes, only seen classes, and only unseen classes.}
\label{fig:plot_awa2_25x5x5c}
\end{figure}

\textbf{Model Evolution.~~} To showcase how the models evolve, plots contrasting the proposed methods with each base model (PODNet and UCIR) task by task appear in \autoref{fig:plot_awa2_25x5x5c}. The plots show how, on early tasks, the main advantage of the proposed model is its ability to guess on the future classes, while on the final task no future classes remain, but the proposed model still keeps a (more modest, but definitive) advantage.

\begin{table}
\caption{Further experiments where the initial task size correspond to standard zero-shot seen classes \cite{xian2019awa2}. We report Continual and Final Accuracies for PODNet on AwA2 and aP\&Y.}
\label{tab:initial_seen}
\centering
\begin{tabular}{@{}l|cc|cc@{}}
\toprule
& \multicolumn{2}{c}{AwA2} & \multicolumn{2}{c}{aP\&Y}\\
 &  \multicolumn{2}{c}{40 classes + 5 $\times$ 2 classes} &  \multicolumn{2}{c}{20 classes + 6 $\times$ 2 classes} \\

& Continual & Final & Continual & Final\\
\midrule
PODNet \cite{douillard2020podnet} & 82.84 $\pm$ 0.10 & 84.70 $\pm$ 0.10 & 67.57 \std 0.41 & 65.23 \std 0.50\\
\tableindent+ $\mcL^\text{\tiny{nca-ghost}}$ & 84.99 $\pm$ 0.17 & 86.57 $\pm$ 0.49 & 68.80 \std 0.98 & 67.93 \std 1.24\\
\tableindent+ $\mcL^\text{\tiny{nca-ghost}}$ + $\mcL^\text{\tiny{svm-reg}}$ & 84.47 \std 0.15 & 85.73 \std 0.40 & 69.02 \std 0.46 & 67.97 \std 0.60\\
\bottomrule
\end{tabular}
\end{table}

\textbf{Zero-shot-like Initial Task Setting.~~} This set of experiments (\autoref{tab:initial_seen}) is intended for comparison with zero-shot learning benchmarks \cite{xian2019awa2}, which always use the same split of seen/unseen classes for a given dataset. Our first task in the continual learning contains the classes defined in zero-shot benchmark as seen, and we learn next, in small increment, the remaining classes, i.e., those defined in the zero-shot benchmark as unseen. Because the initial task is larger than previously, fewer future classes remain, and the base models have better performance. Still, the proposed method improves both base methods in both datasets significantly. The setting proposed is different than the — markedly less challenging — setting appearing in Kankuekul et al.\cite{kankuekul2012onlineincrementalzeroshot} and Xue et al.\cite{xue2017incrementalzeroshot}, where the set of unseen classes is fixed, and only more seen classes are added incrementally, without any sample limitations given by rehearsal memory.


\begin{table*}
\caption{Comparison of generated ghost features vs. actual features extracted from future classes'\,samples with PODNet on AwA2 and aP\&Y.}
\label{tab:generated_vs_real}
\centering
\begin{tabular}{@{}l|cc|cc@{}}
 \toprule
& \multicolumn{2}{c}{AwA2}            & \multicolumn{2}{c}{aP\&Y }        \\
 &  \multicolumn{2}{c}{25 classes + 5 $\times$ 5 classes} &  \multicolumn{2}{c}{16 classes + 8 $\times$ 2 classes} \\
  & Continual & Final & Continual & Final\\
\midrule
Our model   & 68.46 \std 0.47 & 79.08 \std 0.53 & 62.73 \std 0.60 & 63.30 \std 0.98\\
\tableindent w/ real features  & 67.65 \std 0.50 & 78.83 \std 0.31 & 61.88 \std 0.52 & 61.70 \std 0.26\\
\gray{Partial oracle}  & \gray{72.94 \std 0.25} & \gray{84.60 \std 0.28} & \gray{63.81 \std 0.29} & \gray{68.03 \std 1.42}\\
\gray{Full oracle}  & --- & \gray{95.40 \std 0.02} & --- & \gray{97.40 \std 0.30}\\
\bottomrule
\end{tabular}
\end{table*}

\textbf{Generator Validation.~~} The generator approximates the feature extractor for the unseen future classes. To validate its effectiveness, we replace the generated features by the actual features from the future images. This form of “cheating”, of course, is not possible in \textit{actual} real-world scenarios, but serves as a metric. \autoref{tab:generated_vs_real} shows the comparison, with the surprising result that generated features performed better than the actual features from samples (respectively first and second row). Note that the latter are extracted once per task. We hypothesize it explains the score difference because the features extractor was never adapted for the unseen classes distribution.
The “oracle” experiments in the third and fourth row in \autoref{tab:generated_vs_real} establish an upper bound for what we could achieve by ”cheating” around the experimental protocol restrictions. The partial oracle from third row is the same model as the second row, fine-tuning the feature extractor with samples coming from the future. The full oracle of the fourth line uses all images from all classes unrestrictedly in a single task. Despite the partial oracle had full access to real future data, we stress that our model's performance with generated future data is close to this upper bound.

\section{Conclusion}

In this work, we introduced prescient continual learning, a new challenging setting for continual learning, where the model trains on a sequence of tasks, each introducing new classes, but has access to prior information about the classes. Although we give the model awareness of future classes, we also test it on all classes: past, present, and future, resulting in a more challenging setting. We proposed ghost model, first of its kind, which uses the paradigm of representation learning to incorporate capabilities of zero-shot learning into the continual learning model in a seamless way. We refined that model with a novel SVM-based regularization loss acting over the feature space to reinforce exclusion zones, reserved for future classes. Finally, we established, in extensive quantitative and qualitative experiments, the advantage of the proposed model over two base models.

We extend continual learning to a novel setting, which may have important impacts on real-life applications. Robots, for example, in addition to learning incrementally, may have to make decisions for concepts/classes for which training data is not yet available. Our work opens a perspective for those challenging scenarios, which may be interpreted as teaching machine models ``\textit{improvisation}''. The risks of providing this kind of freedom to machine models are still unknown and would be a significant issue as the ability of such models improves.

\section{Supplementary material}

\subsection{Generator details}

Our generator has two inputs: a noise vector and the class attributes. The former, of length 15, is the diversity factor of the generation and is sampled from a Gaussian distribution (although a uniform distribution worked as well). The latter is a vector of attributes belonging to the class, with discrete attributes (i.e., is a reptile? can fly? etc.), with binary values for each sample, which we average per class to produce continuous-valued attribute vectors. We keep the attributes fixed, with no tuning during training.

We train the generator and the main model (feature extractor and classifier) alternately, as shown in \autoref{fig:training_procedure}. We always train the generator at the end of the task, after the feature extractor has adapted to the new distribution. We then use the generator to produce ghost samples for the next task (unless we have reached the last task, with no unseen classes).

When training the generator, we first extract features for all seen classes, with free access to the training samples for the present classes, but only a limited number (given by the rehearsal memory) for the past classes. For better numerical behavior, we scale each dimension of the extracted features to the interval [0, 1] before feeding them to the generator (and then re-scale the output of the generator back to the original intervals before feeding its samples to the classifier). The generator is trained to minimize the Maximum Mean Discrepancy (MMD) between the features of seen concepts $\vh_c = f^t(\vx_c) \, , \forall c \in C^{1:t}$ and the representation it generates $\tilde{\vh}_c = g^t(\mathbf{\xi}, \mathbf{E}_{c}) \,, \forall c \in C^{1:t}$:
\begin{equation}
    \mcL^\text{\tiny{MMD}}_{\Theta_g} = \left\Vert \frac{1}{N} \sum_{i=1}^N \phi\big(\vh^{(i)}_c\big) - \frac{1}{N} \sum_{j=1}^N \phi\big(\tilde{\vh}^{(j)}_c\big) \right\Vert^2\,, \quad c \in C^{1:t}\,,
    \label{eq:mmd_loss}
\end{equation}
with $\phi(\cdot)$ being a Gaussian kernel. The trained generator uses the attributes of the classes — which is the only information we have about them — to estimate sample features that we call \textit{ghost features}. To better estimate the statistics, we use all the real features of a single seen class per batch. We denote the amount of real features by $N$. The generator produces as much ghost features as real features.


Our main model has mechanisms to fight Catastrophic Forgetting, which we found were sufficient also to protect the generator. The feature extractor has an explicit distillation loss to prevent the problem, and since its output is used to train the generator, the latter is also protected.

\subsection{Overview of the Ghost model}

\begin{figure}
\centering
\includegraphics[width=\linewidth]{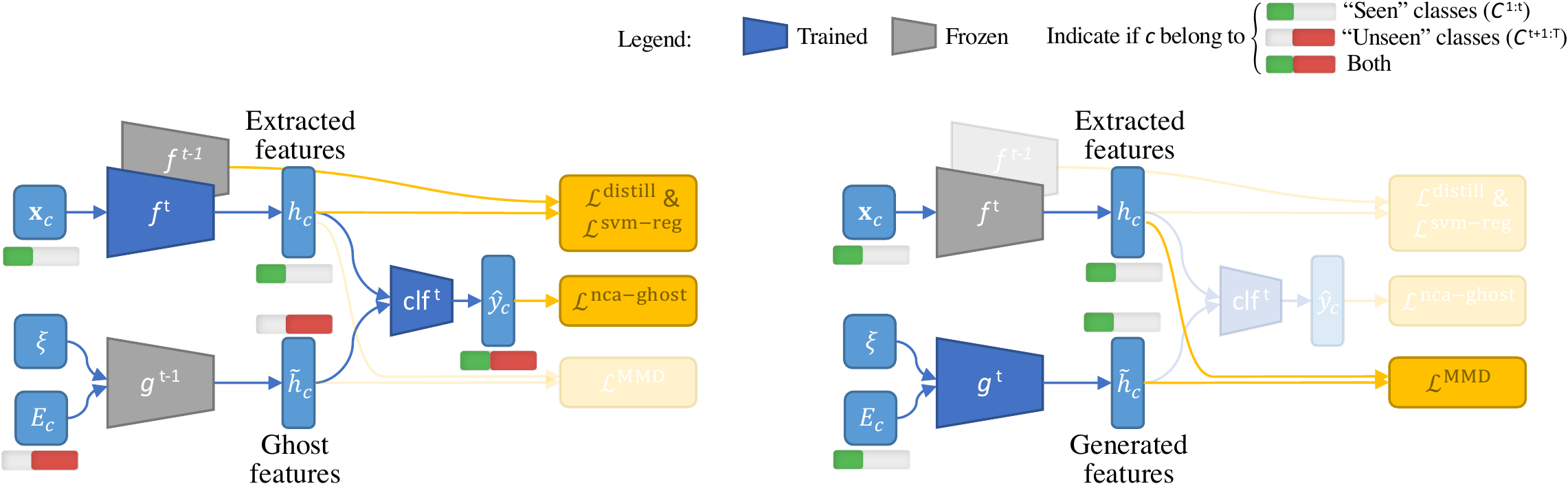}
\begin{subfigure}{.5\textwidth}
  \vspace{2mm}
  \centering
  \caption{Training of the classifier}
  \label{fig:training_cls}
\end{subfigure}%
\begin{subfigure}{.5\textwidth}
  \vspace{2mm}
  \centering
  \caption{Training of the generator}
  \label{fig:training_gen}
\end{subfigure}
\caption{Procedure to train our model applied at each task/step: (a) a complete classifier is learned with seen and unseen features ($\mcL^\text{\tiny{nca-ghost}}$). The feature extractor is protected from catastrophic forgetting ($\mcL^\text{\tiny{distill}}$), and constrained to separate seen classes from unseen/ghosts classes ($\mcL^\text{\tiny{svm-reg}}$). (b) Once a task is done, the generator is fine-tuned on the new latent space ($\mcL^\text{\tiny{MMD}}$) on seen classes. Notice that for the first and last tasks, the classifier does not use the ghost features.}
\label{fig:training_procedure}
\end{figure}

In addition to the description in the main paper, we propose here two additional views of the ghost model. In \autoref{fig:training_procedure}, we show a schematic view of the model, highlighting which components are learned in which turn. Remark that the main model (feature extractor and classifier) is trained alternately with the generator: while one is trained, the other is kept frozen.

\autoref{algo:task_training} is the algorithm of our model in pseudo-code, showcasing a procedural view of the execution of one task.

\begin{algorithm}
\caption{Task procedure of the Ghost model}
\label{algo:task_training}
\begin{algorithmic}[1]
 \REQUIRE task id $t$
 \REQUIRE $f^t$, $c^t$, and $g^t$
 \REQUIRE Data from new task and rehearsal memory $(X, y)$
 \IF{$t = 1$ or $t = T$}
    \STATE Train $f^t$ and $g^t$ with $\mcL^\text{\tiny{nca}}$ and $\mcL^\text{\tiny{distill}}$ on $(X, y)$ and Ghost samples.
 \ELSE
    \STATE Train $f^t$ and $g^t$ with $\mcL^\text{\tiny{nca-ghost}}$, $\mcL^\text{\tiny{svm-reg}}$ and $\mcL^\text{\tiny{distill}}$ on $(X, y)$ and Ghost samples.
 \ENDIF

 \STATE Evaluate on all classes: $C^{1:t} \cup C^{t+1:T}$.
 
 \IF{$t \ne 1$ and $t < T - 1$}
    \STATE Train $g^t$ with $\mcL^\text{\tiny{MMD}}$ using attributes of seen classes $\mathbf{E}_{c} \,, \forall c \in C^{1:t}$
    \STATE Generate Ghost samples for next task using attributes of unseen classes $\mathbf{E}_{c} \,, \forall c \in C^{t+2:T}$
 \ENDIF
\end{algorithmic}
\end{algorithm}

\subsection{Further pictorial experiments}

\begin{figure}[httb!]
\centering
  \includegraphics[width=0.8\linewidth]{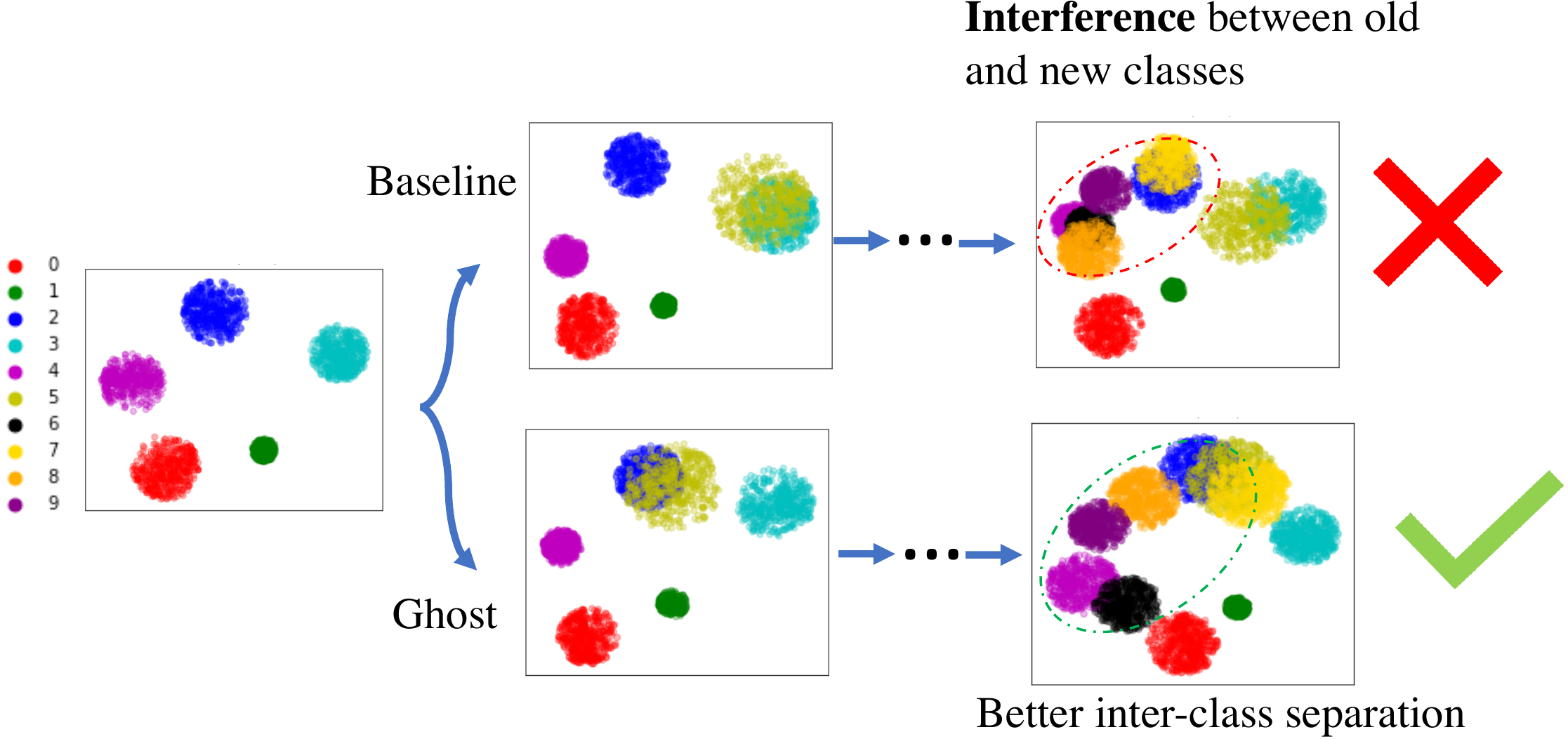}
\caption{Small-scale experiment on MNIST with 6 tasks with a feature space of only two dimensions. We only display, for both PODNET and our ghost model, the first, second, and last task.}
\label{fig:toy_ghost_weights_6steps}
\end{figure}

In the main paper, we evaluate our model on a 3-task split of MNIST. We reproduce here the experiments with a longer continual setting, with 6 tasks. The initial step has half the classes (5), then we incrementally add 5 tasks with 1 new class each. \autoref{fig:toy_ghost_weights_6steps} shows the first, second, and last task for both the base and ghost models. In the base model's last task, we clearly see a considerable overlap among incoming classes (denoted by a red circle). That overlap is less marked, almost non-existent, in the ghost model (denoted by a large green circle). Interferences among classes are minimized.

\begin{figure}
\centering
\begin{subfigure}{.33\textwidth}
  \centering
  \includegraphics[width=\linewidth]{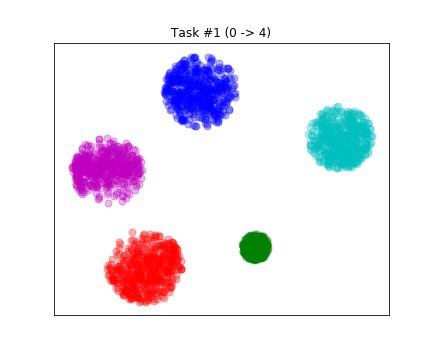}
  \caption{First task}
  \label{fig:toy_6steps_base_1}
\end{subfigure}%
\begin{subfigure}{.33\textwidth}
  \centering
  \includegraphics[width=\linewidth]{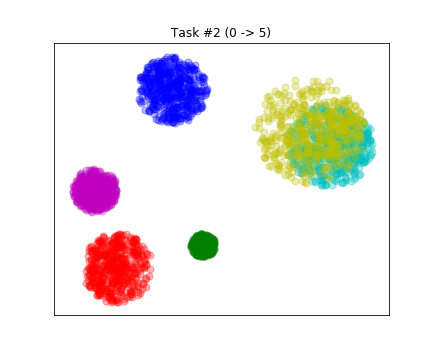}
  \caption{Second task}
  \label{fig:toy_6steps_base_2}
\end{subfigure}
\begin{subfigure}{.33\textwidth}
  \centering
  \includegraphics[width=\linewidth]{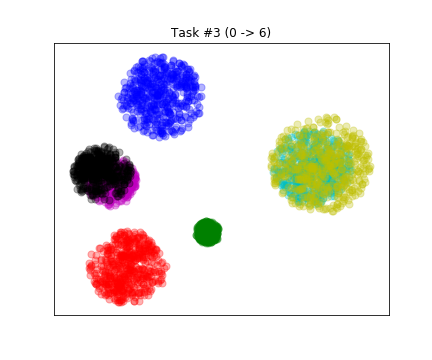}
  \caption{Third task}
  \label{fig:toy_6steps_base_3}
\end{subfigure}

\begin{subfigure}{.33\textwidth}
  \centering
  \includegraphics[width=\linewidth]{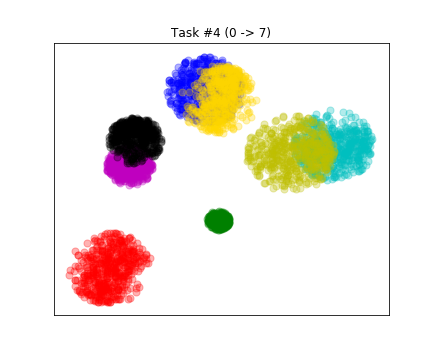}
  \caption{Fourth task}
  \label{fig:toy_6steps_base_4}
\end{subfigure}
\begin{subfigure}{.33\textwidth}
  \centering
  \includegraphics[width=\linewidth]{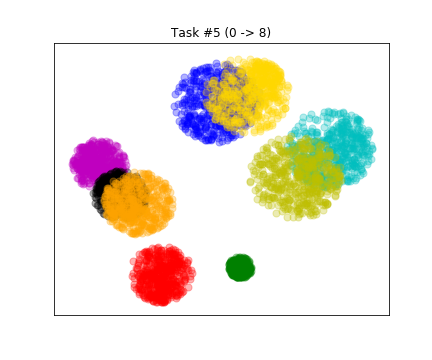}
  \caption{Fifth task}
  \label{fig:toy_6steps_base_5}
\end{subfigure}
\begin{subfigure}{.30\textwidth}
  \centering
  \includegraphics[width=\linewidth]{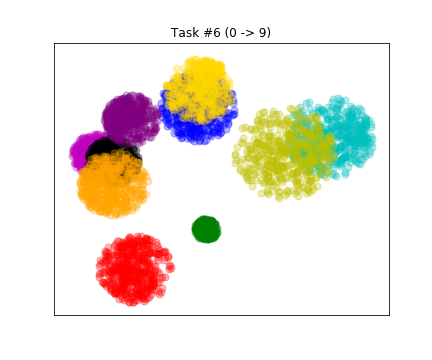}
  \caption{Sixth and last task}
  \label{fig:toy_6steps_base_6}
\end{subfigure}

\caption{Small-scale PODNet on a 6-steps split MNIST, with latent space of dimension two.}
\label{fig:toy_6steps_base}
\end{figure}
\begin{figure}
\centering
\begin{subfigure}{.33\textwidth}
  \centering
  \includegraphics[width=\linewidth]{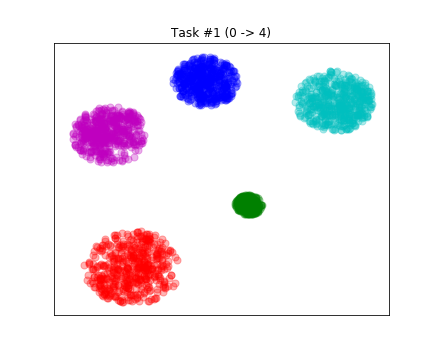}
  \caption{First task}
  \label{fig:toy_6steps_ghost_1}
\end{subfigure}%
\begin{subfigure}{.33\textwidth}
  \centering
  \includegraphics[width=\linewidth]{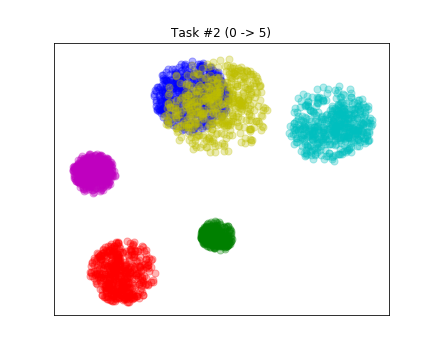}
  \caption{Second task}
  \label{fig:toy_6steps_ghost_2}
\end{subfigure}
\begin{subfigure}{.33\textwidth}
  \centering
  \includegraphics[width=\linewidth]{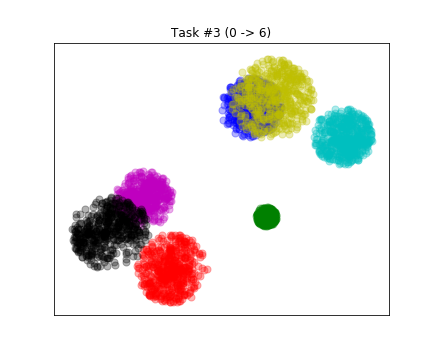}
  \caption{Third task}
  \label{fig:toy_6steps_ghost_3}
\end{subfigure}

\begin{subfigure}{.33\textwidth}
  \centering
  \includegraphics[width=\linewidth]{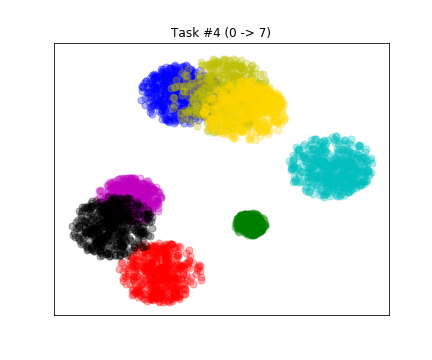}
  \caption{Fourth task}
  \label{fig:toy_6steps_ghost_4}
\end{subfigure}
\begin{subfigure}{.33\textwidth}
  \centering
  \includegraphics[width=\linewidth]{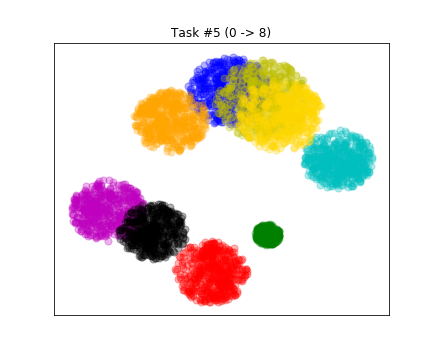}
  \caption{Fifth task}
  \label{fig:toy_6steps_ghost_5}
\end{subfigure}
\begin{subfigure}{.30\textwidth}
  \centering
  \includegraphics[width=\linewidth]{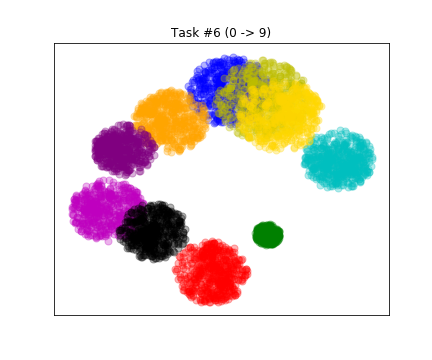}
  \caption{Sixth and last task}
  \label{fig:toy_6steps_ghost_6}
\end{subfigure}

\caption{Small-scale Ghost model on a 6-steps split MNIST, with latent space of dimension two.}
\label{fig:toy_6steps_ghost}
\end{figure}

\autoref{fig:toy_6steps_base} and \autoref{fig:toy_6steps_ghost} show all intermediate tasks for respectively the base model and the ghost model on 6-tasks split MNIST. 

\subsection{Ghost visualization}

\begin{figure}[httb!]
\centering
\begin{subfigure}{.45\textwidth}
  \centering
  \includegraphics[width=\linewidth]{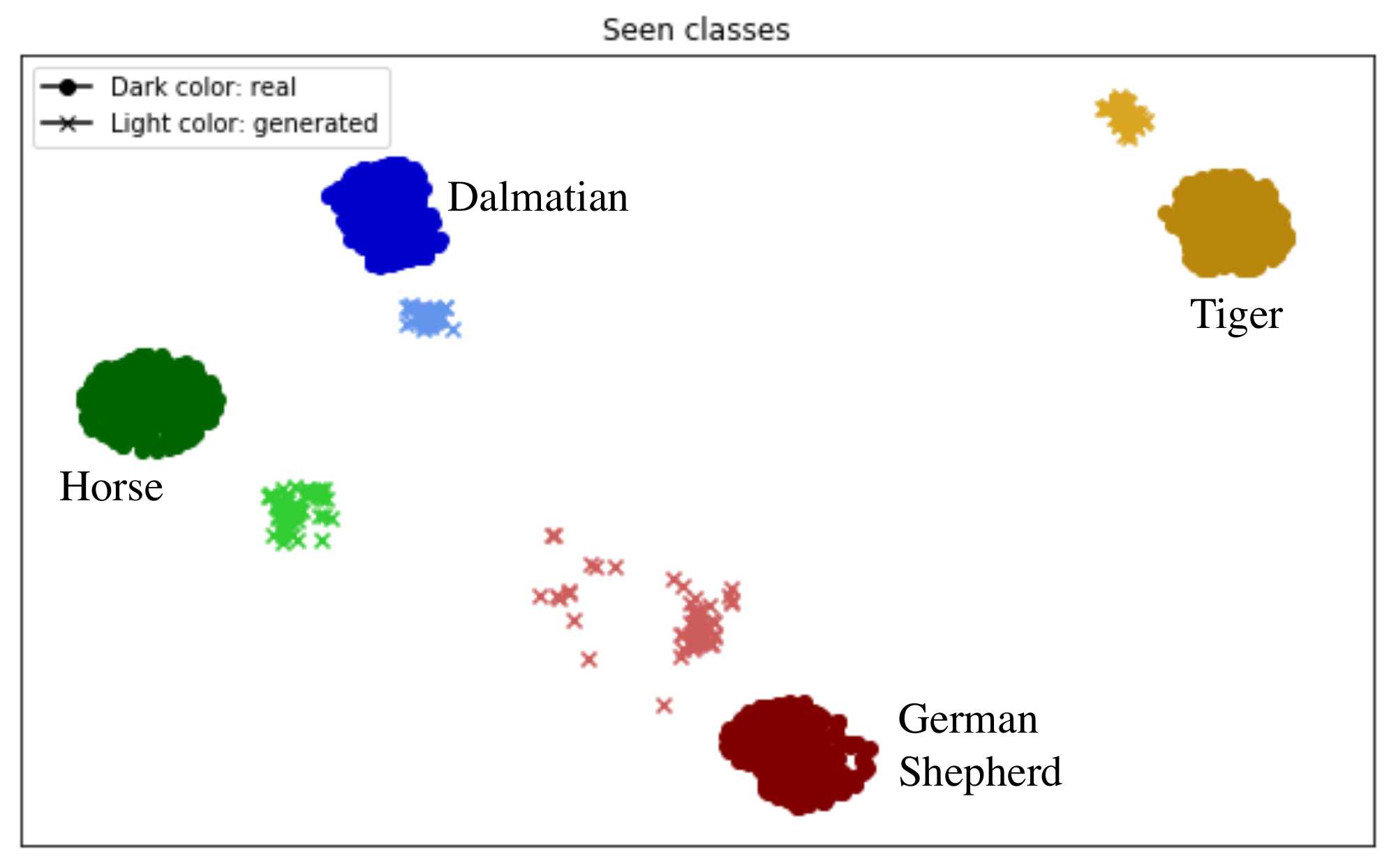}
  \caption{Generator interpolation for \textit{seen} classes.}
  \label{fig:tsne_seen}
\end{subfigure}%
\begin{subfigure}{.45\textwidth}
  \centering
  \includegraphics[width=\linewidth]{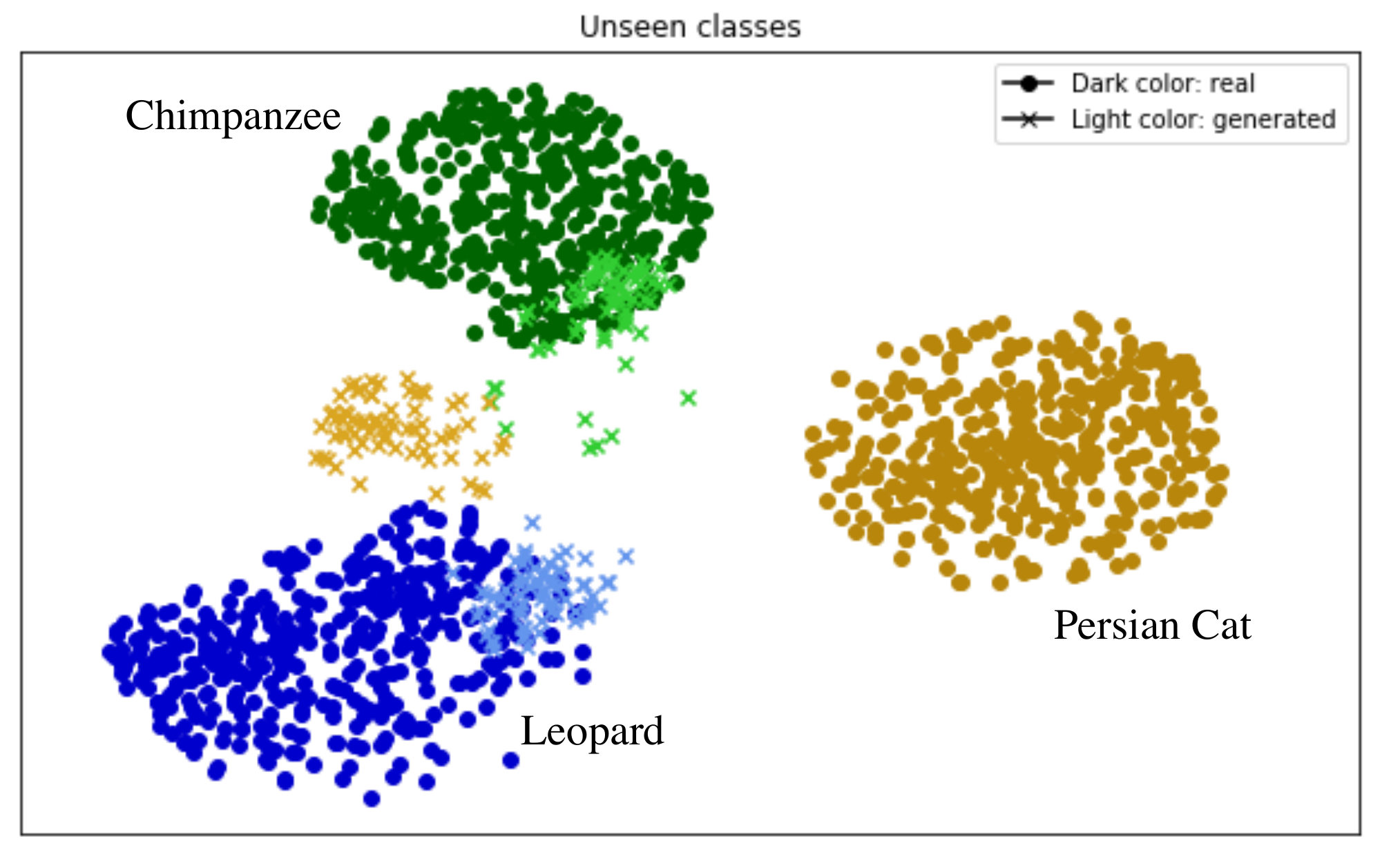}
  \caption{Generator extrapolation for \textit{unseen} classes.}
  \label{fig:tsne_unseen}
\end{subfigure}
\caption{t-SNE of the latent space. Dark colors indicate real features, while lighter colors denote their generated homologous. Real features extracted with $f^t$, and ghost features sampled from $g^t$. Generation in (a) is both well-located and tightly bound because the GMMN was trained to approximate those seen classes. In (b), the generator is asked to extrapolate to unseen classes only from their attributes, resulting in more spread features — still surprisingly, in general, well-located. Notice that the even the real features in (b) gets more spread, since the feature extractor was never trained on those classes.}
\label{fig:tsne_projection_ghost}
\end{figure}

We inspect the ghost samples sampled from our generator with visualization in \autoref{fig:tsne_projection_ghost}: we compare the features extracted from the actual images with their generated homologous, produced by the GMMN on the AwA2 dataset. To allow visualization, we reduce the dimensionality of the features to two with t-SNE. In \autoref{fig:tsne_seen}, we compare, at task $t$, the actual features (in dark colors) to generated features (in light colors) for seen classes. The generated features are, in most cases, near the clusters of real features, and all clusters — real and generated are tightly bound. We then compare in \autoref{fig:tsne_unseen} the real and generated features on unseen classes. Compared to \autoref{fig:tsne_seen}, even the real features show a bigger spread, due, we believe, to the feature extractor $f^t$ not having met those unseen images. The generated features are also more spread but are still, in general, reasonably placed, indicating the ability of the generator to extrapolate — at least partially — the features for unseen classes from their attributes.

\subsection{Overhead of SVMs training}

Training the SVMs for $\mcL^{\text{\tiny{svm-reg}}}$ introduces a computational overhead. To minimize it, we limit the number of features per class to 500. Moreover, as we advance towards later tasks, fewer unseen classes remain, and thus we have fewer SVMs to train. Overall, an experiment on AwA2, with our setting of 25 classes + 5 $\times$ 5 classes, takes 5 hours to train. We observed that our SVM-based regularization extends that time by less than 5 minutes on average, an overhead of less than 2\%, which we deemed acceptable. For reference, the SVMs were trained on a machine with 10 CPU cores of 3.90GHz each.

\subsection{Implementation Details} For all datasets and settings, we set the classification margin $\delta=0.6$, and the SVM latent-space regularization additional margin $\tau=1$. We train the feature-extractor-and-classifier pipeline for 90 epochs with an SGD optimizer, learning rate of 0.1, cosine scheduling, and weight decay of $10^{-4}$. We train the generator for 1200 epochs, with an Adam optimizer and a learning rate of $10^{-5}$. Finally, following \cite{hou2019ucir,douillard2020podnet}, we fine-tune the classifier for 60 epochs (with the feature extractor frozen and a small learning rate of $10^-4$) at the end of every task (except the last one). We found useful to balance the bias towards the seen classes against the unseen classes. With the POD distillation \cite{douillard2020podnet}, we set $\lambda_1=3$  for AwA2, and $\lambda_1=15$ for aP\&Y; with the Less-Forget distillation \cite{hou2019ucir}, we set $\lambda_1=4$ for both datasets. We always set $\lambda_2=10^{-3}$, moreover we apply it on L2-normalized features. Finally, we do not reinitialize the models between tasks: $f^t$ results from training $f^{t-1}$ on task $t$, etc. On the rehearsal memory limitation, we follow the strict setting of Hou et al.~\cite{hou2019ucir}, keeping only $s=20$ training images per past class.

\subsection{Datasets details}

We train our model on three datasets: MNIST, AwA2, and aP\&Y. Baselines and our Ghost models are run on the exact same data/class splits, with the exact same preprocessing.

\textbf{MNIST} This dataset has ten classes: handwritten digits ranging from '0'' to '9'. It has a training set of 60,000 images and a test set of 10,000 images. We used for validation set, a subset of 10,000 examples of the training set. Images are in black\&white (one channel) and of dimension $28\times28$. We convert the pixels values to the range [0, 1] and then normalize by the mean and standard deviation of the training dataset.

\textbf{AwA2} This dataset has 50 animals classes. It has a training set of 29,857 images and a test set of 7,465 images. We used for validation set a subset of 8,000 images of the training set. Images are in RGB color. We convert the pixel values to the range [0, 1] and normalize by the mean and standard deviation of the training dataset. Train images are randomly cropped to a square of $224\times224$ and are randomly flipped horizontally. Test images are resized to $256\times256$ and then center cropped to $224\times224$. 

\textbf{aP\&Y} This dataset has 32 classes of everyday objects. It has a training set of 12,269 images and a test set of 3,068 images. We used for validation set a subset of 4,000 images of the training set. Images are in RGB color. We convert the pixel values to the range [0, 1] and normalize by the mean and standard deviation of the training dataset. Train images are randomly cropped to a square of $224\times224$ and are randomly flipped horizontally. Test images are resized to $256\times256$ and then center cropped to $224\times224$.

\subsection{Reproducibility}

\textbf{Code Dependencies} The Python version is  3.7.6. We used the PyTorch (version 1.2.0) deep learning framework and the libraries Torchvision (version 0.4.0), NumPy (version 1.17.2), Pillow (version 6.2.1), and Matplotlib (version 3.1.0). The CUDA version is 10.2.

We will release the code publicly at the paper acceptance.

\textbf{Hardware \& Training duration} We ran our experiments on 3 Titan Xp GPUs with 12 Go of VRAM each. Each experiment had access to 10 CPU cores of 3.90 GHz each, and used at most 3 Go of RAM and 8 Go of VRAM. A single experiment run on AwA2 took on average 5 hours and, on aP\&Y, 3 hours. We ran each experiment thrice with different random seeds (1, 2, and 3).

\bibliographystyle{plainnat}
\bibliography{egbib}

\end{document}